\documentclass{article}

    \PassOptionsToPackage{numbers, compress}{natbib}

\usepackage[preprint]{neurips_data_2023}

\usepackage[utf8]{inputenc} 
\usepackage[T1]{fontenc}    
\usepackage{hyperref}       
\usepackage{url}            
\usepackage{booktabs}       
\usepackage{amsfonts}       
\usepackage{nicefrac}       
\usepackage{microtype}      
\usepackage{xcolor}         
\usepackage{multirow}
\usepackage{pifont}
\usepackage{graphicx}
\usepackage{caption}
\usepackage{wrapfig}

\title{OCTScenes: A Versatile Real-World Dataset of Tabletop Scenes for Object-Centric Learning}

\author{%
Yinxuan Huang \And Tonglin Chen \And Zhimeng Shen \AND Jinghao Huang \And Bin Li$^{*}$ \And Xiangyang Xue\thanks{Corresponding author.} \\
\AND
  School of Computer Science, Fudan University \\
  \texttt{\{yxhuang22, zmshen22, jhhuang22\}@m.fudan.edu.cn} \\
  \texttt{\{tlchen18, libin, xyxue\}@fudan.edu.cn} \\
}

\begin{document}

\maketitle

\begin{abstract}
  Humans possess the cognitive ability to comprehend scenes in a compositional manner. To empower AI systems with similar capabilities, object-centric learning aims to acquire representations of individual objects from visual scenes without any supervision. Although recent advances in object-centric learning have made remarkable progress on complex synthesis datasets, there is a huge challenge for application to complex real-world scenes. One of the essential reasons is the scarcity of real-world datasets specifically tailored to object-centric learning. To address this problem, we propose a versatile real-world dataset of tabletop scenes for object-centric learning called OCTScenes, which is meticulously designed to serve as a benchmark for comparing, evaluating, and analyzing object-centric learning methods. OCTScenes contains 5000 tabletop scenes with a total of 15 objects. Each scene is captured in 60 frames covering a 360-degree perspective. Consequently, OCTScenes is a versatile benchmark dataset that can simultaneously satisfy the evaluation of object-centric learning methods based on single-image, video, and multi-view. Extensive experiments of representative object-centric learning methods are conducted on OCTScenes. The results demonstrate the shortcomings of state-of-the-art methods for learning meaningful representations from real-world data, despite their impressive performance on complex synthesis datasets. Furthermore, OCTScenes can serve as a catalyst for the advancement of existing methods, inspiring them to adapt to real-world scenes. Dataset and code are available at \url{https://huggingface.co/datasets/Yinxuan/OCTScenes}.
\end{abstract}

\section{Introduction}
Scene comprehension is one of the fundamental tasks of computer vision. Compared with directly perceiving the whole scene, perceiving the scene compositionally makes it easier for the model to acquire relevant knowledge and understand the scene better~\cite{fodor1988connectionism}. Object-centric learning methods aim to learn compositional scene representations in an unsupervised manner~\cite{10151910}, which are better applicable to downstream tasks. Existing object-centric learning methods~\cite{eslami2016attend, engelcke2019genesis, engelcke2021genesis, lin2019space, yuan2022unsupervised, emami2021efficient, kipf2022conditional, singh2022illiterate, kabra2021simone} have achieved remarkable performance on synthetic scene datasets, but their application to real-world scenes remains a significant challenge. An important reason is that no real-world scene datasets are tailored for object-centric learning methods.

Recently, many research works of object-centric learning have been successful on a variety of synthetic datasets that have been made available over the past few years, such as CLEVR~\cite{johnson2017clevr}, SHOP-VRB~\cite{nazarczuk2020shop}, and MOVi~\cite{greff2022kubric}. Although some real-world datasets have been used to evaluate object-centric learning methods, they are unsuitable benchmarks for object-centric learning. For example, the Sketchy~\cite{cabi2019scaling} dataset used in GENESIS-V2~\cite{engelcke2021genesis} lacks the ground truth object-level masks for measuring segmentation performance. The Weizmann Horse~\cite{borenstein2004learning} and APC~\cite{zeng2017multi} datasets used in MarioNette~\cite{smirnov2021marionette} and GENESIS-V2~\cite{engelcke2021genesis} only include one foreground object in the scene. Besides, the real-world datasets used for semantic segmentation or instance segmentation, such as COCO~\cite{lin2014microsoft} and VOC~\cite{everingham2010pascal}, may have no real sense of background and the mask may contain complex objects (for example, COCO may treat the whole person as an object). These datasets bring a massive challenge for unsupervised object-centric learning methods as these methods may segment complex objects into several parts as humans do. The lack of real-world datasets challenges innovative research of object-centric learning.

To address this limitation, we propose OCTScenes, a versatile real-world dataset of tabletop scenes designed explicitly for object-centric learning. OCTScenes contains 5,000 scenes, each consisting of 60 images captured from different viewpoints. As a result, OCTScenes is suitable for various object-centric learning models including single-image-based, video-based, and multi-view based. In OCTScenes, the scenes are set on a table placed on the floor, with objects randomly selected and manually placed on the table. To capture the dataset, we employed a robot equipped with a 3D camera. The robot's movement followed a predefined circular path determined by the algorithm, enabling it to capture images from a 360-degree rotation around the scene. The 3D camera used in the dataset captures both color and depth information, resulting in RGB-D images for each scene.

In experiments, we utilize the OCTScenes dataset to benchmark various representative or state-of-the-art object-centric learning methods based on single-image, video, and multi-view. Extensive experiments demonstrate that some methods outperforming complex synthetic datasets may work poorly on OCTScenes. It implies that the research of object-centric learning urgently needs a tailor-made real-world scene dataset as a benchmark, and the proposed OCTScenes dataset is essential for developing object-centric learning methods.

In summary, the contributions of this paper are as follows:
\begin{itemize}
\item We present OCTScenes, the first real-world RGB-D dataset specific for object-centric learning. Along with the per-frame raw data, we provide segmentation ground truth for evaluation.
\item OCTScenes is a versatile dataset that is suitable for single-image-based, video-based, and multi-view-based object-centric learning methods.
\item We demonstrate the effectiveness of the dataset in advancing state-of-the-art methods while highlighting the limitations of current methods and datasets in generalizing to the real world.
\end{itemize}

We expect OCTScenes to stimulate the development of novel algorithms. Furthermore, we call for an evaluation and comparison of future work on OCTScenes.

\section{Related work}
\paragraph{Synthesis Datasets for Object-Centric Learning} There are several synthesis datasets available for object-centric learning. Earlier approaches were applied to 2D datasets, such as Shapes~\cite{hubert1985comparing}, MNIST~\cite{lecun1998gradient}, dSprites~\cite{matthey2017dsprites} and Abstract Scene~\cite{zitnick2013bringing}. Nevertheless, these 2D datasets are relatively simple and fail to capture the three-dimensional perspective projection relationships in the real world. Consequently, their limited representation makes it challenging to generalize their findings to real-world applications. Many complex 3D synthesis datasets have been proposed to address this limitation, such as CLEVR~\cite{johnson2017clevr}, SHOP-VRB~\cite{nazarczuk2020shop}, CLEVRTex~\cite{karazija2021clevrtex}, and MOVi~\cite{greff2022kubric}. These datasets are generated using the Blender 3D engine, featuring diverse backgrounds and objects with a wide range of materials, shapes, and colors. While initially introduced as a visual question-answering dataset, CLEVR has become a benchmark for object-centric learning. SHOP-VRB provides scenes with various kitchen objects and appliances. Based on CLEVR, CLEVRTex comprises challenging objects featuring diverse materials that include repeating patterns and small details. In comparison, MOVi contains more realistic objects and backgrounds. The main difference between OCTScenes and the aforementioned datasets is that OCTScenes is a real-world dataset, captured directly from cameras, as opposed to being generated through a rendering engine.

\paragraph{Unsupervised Scene Understanding in Natural Scenes} Many real-world datasets can be used for unsupervised scene understanding, such as PascalVOC~\cite{everingham2010pascal} and COCO~\cite{lin2014microsoft}. However, current object-centric learning models are not yet able to handle the diverse real-world images featured in such datasets, and these datasets mainly focus on outdoor scenes with a background occupying the main image, which is not suitable for learning object-centric representations. Besides, some other real-world scene datasets have been employed, such as Sketchy~\cite{cabi2019scaling}, Weizmann Horse~\cite{borenstein2004learning} and APC~\cite{zeng2017multi}. However, Sketchy lacks the ground truth segmentation masks for evaluation, while Weizmann Horse and APC only contain a single foreground object in the scene. These limitations render them inadequate as benchmarks for evaluating various object-centric learning methods. In contrast, OCTScenes has been specifically designed to overcome these limitations, making it the first real-world dataset that provides multi-object and multi-view scenes for object-centric learning.

In summary, we conduct a comprehensive comparison between our dataset and other commonly used datasets in object-centric learning, which includes both synthetic datasets and real-world datasets. In Table~\ref{tab:overview}, OCTScenes stands out as the only real-world dataset that features multiple objects within each scene and multiple views of the same scene and provides segmentation annotations for evaluation. This unique feature sets OCTScenes apart from other datasets, highlighting its superiority in facilitating object-centric learning.

\begin{table}[ht]
  \caption{Overview of datasets for object-centric learning. Multi-Object refers to whether there are multiple objects present within a scene. Multi-Frame indicates whether the dataset includes multiple frames or views of the same scene. Annotation signifies whether the segmentation maps are provided.}
  \label{tab:overview}
  \centering
  \begin{tabular}{ccccc}
    \toprule
    Dataset & Type & Multi-Object & Multi-Frame & Annotation \\
    \midrule
    Multi-Shapes~\cite{reichert2011hierarchical} & Synthesis, 2D & \ding{51} & \ding{55} & \ding{51} \\
    \midrule
    TexturedMNIST~\cite{greff2016tagger} & Synthesis, 2D & \ding{51} & \ding{55} & \ding{51} \\
    \midrule
    MultiMNIST~\cite{sabour2017dynamic} & Synthesis, 2D & \ding{51} & \ding{55} & \ding{51} \\
    \midrule
    Multi-dSprites~\cite{burgess2019monet} & Synthesis, 2D & \ding{51} & \ding{55} & \ding{51} \\
    \midrule
    Abstract Scene~\cite{zitnick2013bringing} & Synthesis, 2D & \ding{51} & \ding{55} & \ding{51} \\
    \midrule
    CLEVR~\cite{johnson2017clevr} & Synthesis, 3D & \ding{51} & \ding{55} & \ding{51} \\
    \midrule
    SHOP-VRB~\cite{nazarczuk2020shop} & Synthesis, 3D & \ding{51} & \ding{55} & \ding{51} \\
    \midrule
    CLEVRTEX~\cite{karazija2021clevrtex} & Synthesis, 3D & \ding{51} & \ding{55} & \ding{51} \\
    \midrule
    MOVi~\cite{greff2022kubric} & Synthesis, 3D & \ding{51} & \ding{51} & \ding{51} \\
    \midrule
    Sketchy~\cite{cabi2019scaling} & Real-World & \ding{51} & \ding{51} & \ding{55} \\
    \midrule
    Weizmann Horse~\cite{borenstein2004learning} & Real-World & \ding{55} & \ding{55} & \ding{51} \\
    \midrule
    APC~\cite{zeng2017multi} & Real-World & \ding{55} & \ding{51} & \ding{51} \\
    \midrule
    OCTScenes (ours) & Real-World & \ding{51} & \ding{51} & \ding{51} \\
    \bottomrule
  \end{tabular}
\end{table}

\section{OCTScenes}

\subsection{Dataset designment}
We introduce OCTScenes, a real-world multi-functional dataset designed to present the next challenge in unsupervised object-centric learning. It contains multiple scenes with different objects.

\begin{figure}[ht]
  \begin{minipage}[]{.4\linewidth}
    \centering
    \includegraphics[width=\columnwidth]{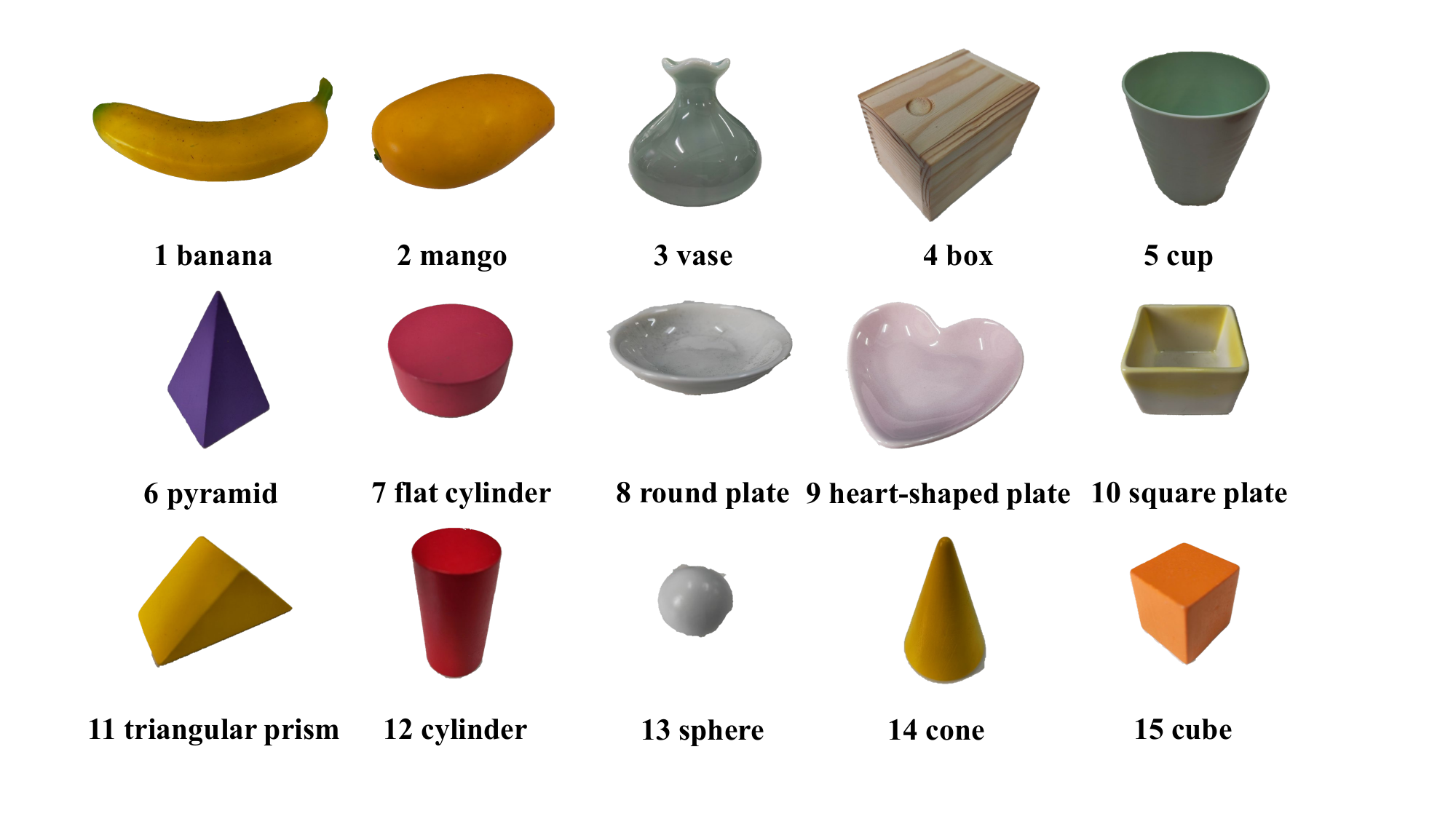}
    \caption{Objects of the dataset.}
    \label{fig:objects}
  \end{minipage}
  \begin{minipage}[]{.6\linewidth}
  \captionof{table}{Dataset configuration}
  \label{tab:dataset}
  \centering
   \scalebox{0.65}{
  \begin{tabular}{c|ccc|ccc}
    \toprule
    Dataset    & \multicolumn{3}{c|}{OCTScenes-A}    &\multicolumn{3}{c}{OCTScenes-B} \\
    \midrule
    Image Size & 640$\times$480 & 256$\times$256 & 128$\times$128 & 640$\times$480 & 256$\times$256 & 128$\times$128 \\
    \midrule
    Spilt   & Train & Valid & Test  & Train & Valid & Test \\
    \midrule
    \# of Scenes    &  3000 & 100 & 100 & 4800 & 100 & 100 \\
    \midrule
    \# of Catergories & \multicolumn{3}{c|}{11} & \multicolumn{3}{c}{15}    \\
    \# of Objects  & \multicolumn{3}{c|}{1$\sim$6} & \multicolumn{3}{c}{1$\sim$10} \\
    \# of Views & \multicolumn{3}{c|}{60} & \multicolumn{3}{c}{60} \\
    \bottomrule
  \end{tabular}}
  \end{minipage}
\end{figure}

\paragraph{Object} Our dataset comprises 15 distinct types of objects, encompassing everyday articles such as a banana, a mango, and a vase, as well as simple geometric shapes like a pyramid, a flat cylinder, and a cylinder. These objects are frequently encountered in our daily routines and display a wide range of characteristics in terms of their shape, color, and materials. Certain objects may present perceptual challenges due to their unique properties. For instance, the vase, which is crafted from ceramic, features a smooth surface that reflects light, thus making it difficult to accurately perceive its true colors. These objects are shown in Figure~\ref{fig:objects}.

\paragraph{Scene} The scene in our dataset consists of a fixed background and multiple foreground objects. The background is a small wooden table placed on the floor and surrounded by baffles, which remains consistent throughout the dataset. We populate the scene with a varying number of objects, ranging from 1 to 10. Each object is randomly selected and manually placed on the table, without any stacking. It is important to note that occlusion between objects is allowed, resulting in scenes where some objects may be completely occluded from a single view, while the 3D geometric relationships between objects can be inferred by analyzing multiple views of the scene. In each scene, both the background and foreground objects remain static, while the images are captured by a camera-equipped robot that moves around the table. This movement introduces variations in viewpoint among the images within a scene. Besides, our dataset exhibits variations in the number, types, positions, and orientations of objects across different scenes. Each scene represents a unique combination, which contributes to the diversity and richness of the dataset.

\paragraph{Dataset} To accommodate diverse research needs, the scenes are divided into two subsets to create datasets with different levels of difficulty. One subset comprises only the first 11 object types, with scenes consisting of 1 to 6 objects, making it comparatively smaller and less complex. The other consists of all 15 object types, with scene compositions ranging from 1 to 10 objects, resulting in a larger and more complex dataset. Furthermore, the former subset is a subset of the latter. More details on these two subsets can be found in Table~\ref{tab:dataset}.

\subsection{Raw data acquisition}

\begin{wrapfigure}{R}{.5\columnwidth}
  \vspace{-0.12in}
  \centering
  \includegraphics[width=.5\columnwidth]{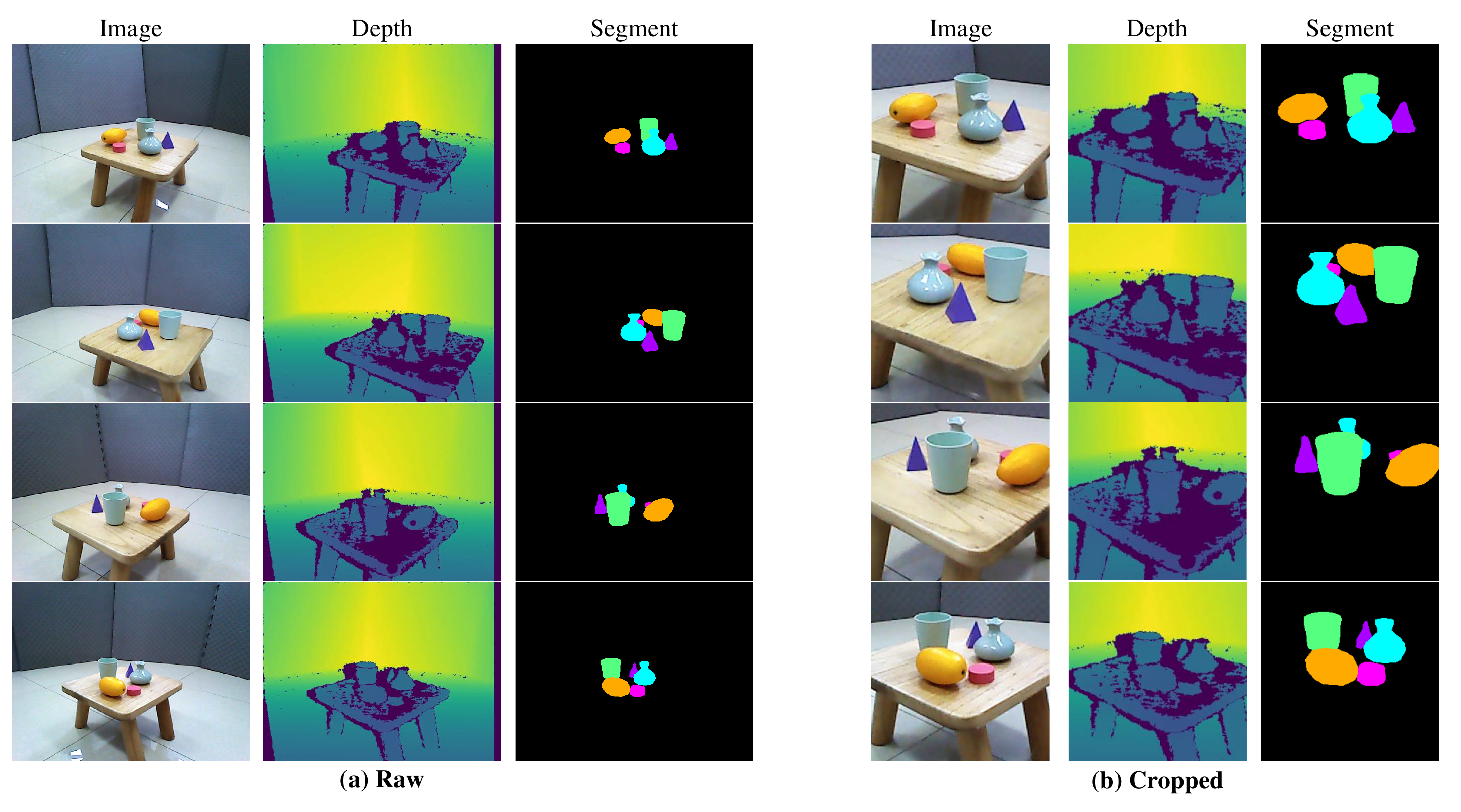}
  \vspace{-0.2in}
  \caption{Examples of images, depth maps, and segmentation maps of the dataset.}
  \label{fig:dataset}
  \vskip -0.1in
\end{wrapfigure}

We employed a three-wheel omnidirectional wheel robot equipped with an Orbbec Astra 3D camera for data collection. The three-wheel omnidirectional wheels are positioned 120 degrees apart, enabling the robot to move effortlessly in all directions. The Orbbec Astra 3D camera operated at a frame rate of 30 frames per second (fps) and captured RGB-D images at a resolution of 640$\times$480. We utilized the official Orbbec SDK\footnote{\url{https://github.com/orbbec/ros_astra_camera}} to write a data collection script that guides the robot along a predefined path around the scene and instructs it to capture RGB-D images while in motion. Throughout the robot's operation, we displayed the perspective of the camera to provide real-time feedback to data collectors, ensuring the quality of the captured images. Figure~\ref{fig:dataset} displays some sample images.

We conducted data collection in a school conference room, executing the data collection script to capture each scene. Throughout the entire data collection process, we attempted to keep the background completely static. However, due to the average duration of 6 hours per day over a total of 18 days, changes in lighting conditions were possible, potentially leading to inconsistent illumination across different scenes.

\subsection{Data processing}

\paragraph{Data cropping and resizing} Since object-centric learning focuses on the foreground objects instead of the background, we generate the bounding boxes of the table to crop the images. We first use LabelImg\footnote{\url{https://github.com/heartexlabs/labelImg}}, a popular image annotation tool, to manually label 1000 images. Each labeled image contains a bounding box that encompasses the entire table along with the objects placed on it. We then train a yolov5\footnote{\url{https://github.com/ultralytics/yolov5}} model, which is a high-performance object detection method, with the labeled data to annotate the remaining dataset. The annotated images are split into 90\% for training and 10\% for validation, achieving a mean Average Precision (mAP) of 0.99 on the validation set. Using the generated annotations, we crop the images by centering them around the midpoint of the corresponding bounding boxes. This process results in cropped images with a resolution of 256$\times$256. Besides, a resized version of images with a resolution of 128$\times$128 is also provided for model training.

\paragraph{Data split} In accordance with the details provided in Table~\ref{tab:dataset}, we partition all scenes into two datasets with different difficulty levels. Each dataset is randomly divided into three subsets: training, validation, and testing. The validation set and testing set consist of 100 scenes each, while the remaining scenes constitute the training set.

\paragraph{Segmentation maps} To evaluate the effectiveness of object-centric learning methods, we generate segmentation maps for the testing set. We first use EISeg\footnote{\url{https://github.com/PaddlePaddle/PaddleSeg/tree/release/2.8/EISeg}}, which is a high-performance interactive automatic annotation tool for image segmentation, to segment the images. Each pixel of the images is annotated into 1 of 16 classes with 15 kinds of objects and 1 background. We manually labeled 6 images of each scene and used the labeled images to train a supervision real-time semantic segmentation model named PP-LiteSeg~\cite{peng2022pp} using the framework PaddleSeg\footnote{\url{https://github.com/PaddlePaddle/PaddleSeg}} to annotation the rest of the data. The annotated images are split into 90\% for training and 10\% validation, achieving a mean Intersection over Union (mIoU) of 0.92 on the validation set.

In summary, we provide three data sizes: raw images (640$\times$480), cropped versions (256$\times$256), and resized versions (128$\times$128). Additionally, we provide two datasets of varying difficulty, each divided into training, validation, and testing sets. For evaluation purposes, the testing set includes segmentation maps.

\section{Models}

Object-centric learning aims to understand a visual scene by parsing the scene into individual objects (or the background) in an unsupervised manner. Existing works have achieved excellent results on complex synthetic datasets, but face a huge challenge in extending them to real-world settings. They can be divided into three categories: single-image-based, video-based, and multi-view-based.

\paragraph{Single-image-based} Object-centric learning was first proposed to learn compositional representations from a single image,  i.e. to extract the object-centric representation for each object or background in the static scenes from only one viewpoint. N-EM~\cite{greff2017neural} and AIR~\cite{eslami2016attend} are two classical compositional scene representation learning methods and are chosen to verify the proposed benchmark. The former obtains the compositional scene representation through iterative updates, and the latter sequentially extracts the representation of each object through rectangular attention. GMIOO~\cite{yuan2019generative} is selected because it combines two ways of inferring object-centric representations in N-EM and AIR, and models the background separately and an infinite number of object scenes for the first time. The methods, including SPACE~\cite{lin2019space}, GENESIS~\cite{engelcke2019genesis}, GENESIS-V2~\cite{engelcke2021genesis}, Slot Attention~\cite{locatello2020object}, EfficientMORL~\cite{emami2021efficient}, SLATE~\cite{singh2022illiterate} and BO-QSA~\cite{Jia2022ImprovingOL}, are chosen as examples of the proposed benchmark because they inspire further research or model the scene in an exciting way for the first time. Specifically, SPACE uses the spatial mixture model to model the background for the first time, with the aim of better learning complex background representations. GENESIS~\cite{engelcke2019genesis} uses the autoregressive model for the first time to model the relationship between objects. GENESIS-V2~\cite{engelcke2021genesis} proposes a differentiable pixel embedding clustering method using the truncated stick process to solve the problem that the model cannot be extended to large images due to the use of RNN for inference. Slot Attention~\cite{locatello2020object} uses a cross-attention mechanism to iteratively update the initialized object-centric representation, which significantly improves the extraction efficiency of scene representations and fosters many powerful variants. EfficientMORL~\cite{emami2021efficient} adds an iterative process to the object representation extracted by Slot Attention to obtain better compositional scene representations. To achieve remarkable results in more complex synthetic scenes, SLATE~\cite{singh2022illiterate} proposed using an autoregressive transformer-based decoder for the first time. BO-QSA~\cite{Jia2022ImprovingOL} proposes learnable queries as slot initializations and improves slot attention with bi-level optimization.

\paragraph{Video-based} Video-based methods take multi-frame videos as input and often exploit temporal cues between adjacent frames to improve performance. SAVi~\cite{kipf2022conditional} is the first method to learn from complex synthetic video, extending Slot Attention from image to video by adding a prediction network that models temporal dynamics and object interactions. Combining the ability of SAVi to exploit temporal information and the power of SLATE to reconstruct complex scenes, STEVE~\cite{singh2022simple} successfully spans the application of object-centric learning from synthetic to complex natural scenes.

\paragraph{Multi-view-based} Multi-view-based methods take multi-view images as input, and scenes are often 360 inward-facing, i.e. objects remain static in the center of the scene while the camera moves around the objects. Unlike video-based methods, multi-view-based methods typically model view representations separately, and the input images of the same scene may be discontinuous in viewpoint. These methods often learn two sets of representations: a set of object representations that are time-invariant, object-level contents of the scene, and a set of view representations that are globally time-varying. SIMONe~\cite{kabra2021simone} and OCLOC~\cite{yuan2022unsupervised} are two representative multi-view-based object-centric learning methods. The former models view representations for the first time and can predict images for the given view representation, the latter can learn view-independent representations of objects by taking images of randomly selected views as input.

\section{Experiments}
In this section, we demonstrate the performance comparison and analysis of the models based on single-image, video, and multi-view on the proposed dataset to verify the versatility of OCTScenes.

\paragraph{Dataset} OCTScenes provides images in three resolutions: 640$\times$480, 256$\times$256, and 128$\times$128. Given that existing object-centric learning methods are generally suitable for 64$\times$64 or 128$\times$128 images as input, we choose the 128$\times$128 image size for all experiments. To make full use of the images for training, we divide each scene (including 60 frames) into 6 sub-scenes with an interval of 10 frames and sample 3 scenes from 6 sub-scenes at intervals. For the OCTScenes-A dataset, the number of training images is 9000$\times$10 and 300$\times$10 for validation and testing. For the OCTScenes-B dataset, the number of training images is 14400$\times$10 and 300$\times$10 for validation and testing.

\paragraph{Metrics} We use an array of precise metrics to evaluate the quality of both segmentation and reconstruction of object-centric learning methods on OCTScenes. We assess segmentation quality with \emph{Adjusted Rand Index} (ARI)~\cite{hubert1985comparing}, \emph{Adjusted Mutual Information} (AMI)~\cite{vinh2009information}, and \emph{mean Intersection over Union} (mIoU). ARI and AMI, which measure the congruence between two data clusters, are robust indicators of superior segmentation performance as their values increase. mIoU, a standard metric for evaluating object segmentation, provides a quantifiable measure of the overlap between the predicted and ground truth segmentation. We further refine our analysis by introducing the terms AMI-A and ARI-A, signifying calculations that consider both the objects and the background, and AMI-O and ARI-O, which focus solely on the objects. We rely on \emph{Minimize Squared Error} (MSE) and \emph{Learned Perceptual Image Patch Similarity} (LPIPS)~\cite{zhang2018perceptual} to evaluate the quality of reconstruction, both of which indicate better reconstruction performance at lower values. The former measures the difference in pixel level and favors blurry results, while the latter measures the difference in feature level and aligns better with human perception.

\paragraph{Implementation details} We used the implementation of N-EM, AIR, SPACE, GENESIS, GENESIS-V2, GMIOO, Slot Attention, and OCLOC from the toolbox of compositional scene representation\footnote{\url{https://github.com/FudanVI/compositional-scene-representation-toolbox}}. The official implementations of EfficientMORL, SLATE, BO-QSA, and STEVE are used. The official implementation of SAVi and the unofficial implementation of SIMONe are modified into a PyTorch version. BO-QSA (mix) is based on a mixture-based decoder, while BO-QSA (trans) is based on a transformer-based decoder. Further implementation details are reported in Appendix~\ref{experiment}.

\subsection{Results}

\begin{figure}[htb]
    \centering
    \includegraphics[width=\columnwidth]{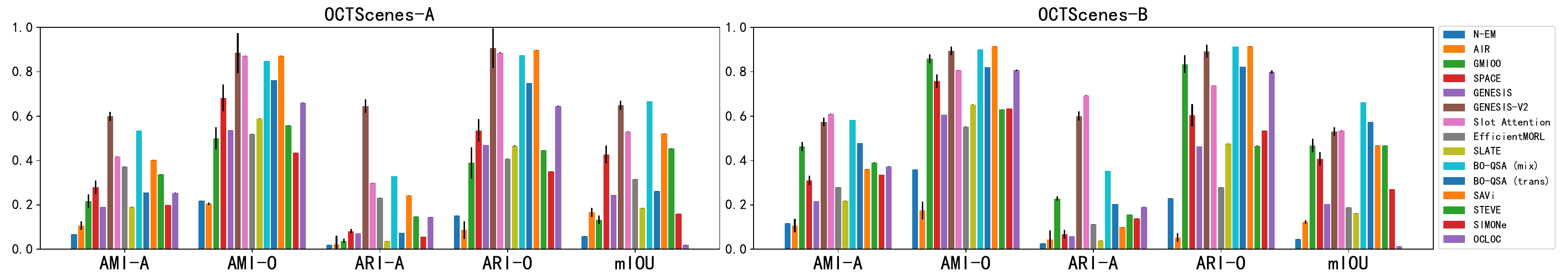}
    \caption{Segmentation performance on OCTScenes-A and OCTScenes-B.}
    \label{fig:seg}
\end{figure}

\begin{figure}[htb]
    \centering
    \includegraphics[width=\columnwidth]{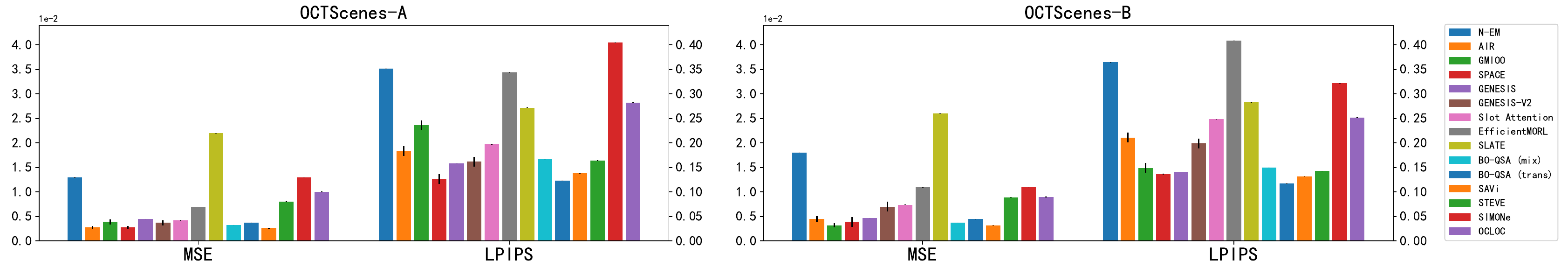}
    \caption{Reconstruction performance on OCTScenes-A and OCTScenes-B. }
    \label{fig:recon}
\end{figure}

We show quantitative experimental results in Table~\ref{tab:da}, Table~\ref{tab:db}, Figure~\ref{fig:seg} and Figure~\ref{fig:recon}. We also visualize qualitative results in Figure~\ref{fig:qua}. More experimental results are presented in Appendix~\ref{results}.

\paragraph{Object segmentation performance} For the methods based on single-image, GENESIS-V2~\cite{engelcke2021genesis}, Slot Attention~\cite{locatello2020object}, and BO-QSA~\cite{Jia2022ImprovingOL} achieve high ARI-O and AMI-O metrics on both OCTScenes-A and OCTScenes-B datasets. Conversely, several earlier methods fail to segment scenes correctly, such as N-EM~\cite{greff2017neural} which splits the whole scene into many fragmented parts, resulting in poor segmentation results. It is worth noting that although SLATE~\cite{singh2022illiterate} performs well on many synthetic datasets, its object segmentation performance on OCTScenes is poor, possibly due to its dependence on the initialization effect of slots in the slot attention module, leading to high randomness and unstable performance. Multiple attempts involving varied random seeds and hyperparameters have failed to yield satisfactory performance. For video-based methods, SAVi~\cite{kipf2022conditional} performs well in object segmentation, while STEVE~\cite{singh2022simple} performs poorly. The reason for its poor performance could be similar to SLATE. When it comes to multi-view-based methods, OCLOC~\cite{yuan2022unsupervised} has better object segmentation performance than SIMONe~\cite{kabra2021simone} on both OCTScenes-A and OCTScenes-B datasets. The visualization results show that OCLOC effectively segments the majority of objects, although it struggles with smaller and obstructed objects. In contrast, SIMONe tends to divide scenes into relatively large clusters, resulting in coarse segmentation results. We attribute this difference to the specific modeling of view in OCLOC, while SIMONe only averages the representations of the same objects in different views to obtain viewpoint information.

\begin{table*}[htb]
  \centering
  \caption{Comparison results of segmentation and reconstruction on OCTScenes-A.}
  \label{tab:da}
  \scalebox{0.64}{
  \begin{tabular}{llccccccc}
    \toprule 
    \bfseries Type & \bfseries Model & \bfseries AMI-A$\uparrow$ & \bfseries AMI-O$\uparrow$ & \bfseries ARI-A$\uparrow$ & \bfseries ARI-O$\uparrow$ & \bfseries mIOU$\uparrow$ & \bfseries MSE $\downarrow$ & \bfseries LPIPS$\downarrow$ \\
    \midrule 
    \multirow{11}{*}{Single-image-based} 
  &N-EM~\cite{greff2017neural}           &     0.068$\pm$3e-4      &     0.219$\pm$9e-4       &    0.020$\pm$2e-4
  &     0.152$\pm$7e-4      &     0.058$\pm$4e-5      &     1.3e-2$\pm$7e-6  &   0.352$\pm$1e-4 \\
  &AIR~\cite{eslami2016attend}            &0.106$\pm$2e-2   &0.205$\pm$5e-3   &0.021$\pm$4e-2
  &0.086$\pm$4e-2      &0.166$\pm$2e-2    & \bfseries 2.8e-3$\pm$3e-4  &0.184$\pm$1e-2  \\ 
  &GMIOO~\cite{yuan2019generative}        &0.217$\pm$3e-2   & 0.501$\pm$5e-2  &0.038$\pm$1e-2
  &0.389$\pm$7e-2      &0.133$\pm$2e-2   &3.9e-3$\pm$5e-4  &0.236$\pm$1e-2   \\ 
  &SPACE~\cite{lin2019space}           &0.279$\pm$3e-2      &0.682$\pm$6e-2   &0.081$\pm$1e-2
  &0.536$\pm$5e-2      &0.427$\pm$4e-2      &\bfseries 2.8e-3$\pm$3e-4  &0.126$\pm$1e-2  \\ 
  &GENESIS~\cite{engelcke2019genesis}         &     0.190$\pm$2e-6      &     0.537$\pm$5e-6       &    0.072$\pm$8e-7
  &     0.470$\pm$3e-6      &     0.246$\pm$2e-6      &     4.5e-3$\pm$2e-8  &  0.158$\pm$4e-6\\ 
  &GENESIS-V2~\cite{engelcke2021genesis}    &\bfseries 0.599$\pm$2e-2   & \bfseries 0.885$\pm$9e-2   & \bfseries 0.645    $\pm$3e-2
  & \bfseries 0.907$\pm$9e-2      & 0.649$\pm$2e-2      &3.7e-3$\pm$5e-4  &0.162$\pm$1e-2  \\ 
  &Slot Attention~\cite{locatello2020object} &0.418$\pm$4e-4   &0.872$\pm$9e-4   &  0.299$\pm$6e-4
  &0.885$\pm$2e-3      &0.531$\pm$6e-4     &4.2e-3$\pm$6e-6 & 0.197$\pm$1e-4 \\ 
  &EfficientMORL~\cite{emami2021efficient}   & 0.373$\pm$4e-4  & 0.519$\pm$9e-4   & 0.232$\pm$3e-4
  & 0.408$\pm$1e-3  & 0.316$\pm$6e-4  &  7.0e-3$\pm$2e-5  & 0.344$\pm$5e-4 \\ 
  &SLATE~\cite{singh2022illiterate}           & 0.190$\pm$2e-4 & 0.590$\pm$9e-4 & 0.037$\pm$2e-4
  & 0.466$\pm$2e-3      & 0.186$\pm$4e-4      & 2.2e-2$\pm$3e-5  & 0.272$\pm$5e-4 \\
  &BO-QSA (mix)~\cite{Jia2022ImprovingOL} & 0.534$\pm$6e-7   &0.849$\pm$2e-6   &  0.330$\pm$2e-7
  &0.875$\pm$1e-6      & \bfseries 0.667$\pm$4e-7     &3.3e-3$\pm$2e-9  & 0.167$\pm$2e-7 \\
  &BO-QSA (trans)~\cite{Jia2022ImprovingOL} &0.255$\pm$4e-7   &0.763$\pm$5e-6   & 0.074$\pm$7e-6
  &0.750$\pm$8e-7      &0.262$\pm$5e-5     &3.7e-3$\pm$3e-6  & \bfseries 0.123$\pm$5e-5 \\
  \midrule
    \multirow{2}{*}{Video-based} 
  &SAVi~\cite{kipf2022conditional}           &\bfseries 0.402$\pm$3e-4 & \bfseries 0.872$\pm$4e-5 & \bfseries 0.242$\pm$1e-4
  & \bfseries 0.899$\pm$4e-4      & \bfseries 0.522$\pm$4e-4      & \bfseries 2.6e-3$\pm$6e-6  & \bfseries 0.138$\pm$1e-4 \\ 
  &STEVE~\cite{singh2022simple}         &     0.337$\pm$3e-4      &     0.559$\pm$6e-4       &    0.147$\pm$7e-5
  &     0.446$\pm$4e-4      &     0.455$\pm$7e-4      &     8.0e-3$\pm$1e-4  &  0.164$\pm$1e-3\\ 
  \midrule
    \multirow{2}{*}{Multi-view-based} 
  &SIMONe~\cite{kabra2021simone}         &     0.200$\pm$1e-5      &     0.436$\pm$5e-5       &    0.056$\pm$3e-6
  &     0.351$\pm$2e-4     &  \bfseries   0.160$\pm$2e-5      &     1.3e-2$\pm$3e-7  &  0.405$\pm$5e-5\\ 
  &OCLOC~\cite{yuan2022unsupervised}          & \bfseries 0.254$\pm$1e-3 & \bfseries 0.661$\pm$1e-3 & \bfseries 0.145$\pm$1e-3
  & \bfseries 0.646$\pm$2e-3 & 0.020$\pm$5e-5 & \bfseries 1.0e-2$\pm$9e-5  & \bfseries 0.282$\pm$1e-3\\  
  \bottomrule 
  
  \end{tabular}
  }
\end{table*}

\begin{table*}[htb]
  \centering
  \caption{Comparison results of segmentation and reconstruction on OCTScenes-B.}
  \label{tab:db}
  \scalebox{0.64}{
  \begin{tabular}{llccccccc}
    \toprule 
    \bfseries Type & \bfseries Model &\bfseries AMI-A$\uparrow$ &\bfseries AMI-O$\uparrow$ &\bfseries ARI-A$\uparrow$ &\bfseries ARI-O$\uparrow$ &\bfseries mIOU$\uparrow$ & \bfseries MSE $\downarrow$ & \bfseries LPIPS $\downarrow$ \\
    \midrule 
    \multirow{11}{*}{Single-image-based} 
  &N-EM~\cite{greff2017neural}           &     0.117$\pm$2e-4      &     0.360$\pm$3e-4       &    0.026$\pm$9e-5
  &     0.230$\pm$4e-4      &     0.046$\pm$6e-5      &     1.8e-2$\pm$5e-6 & 0.365$\pm$8e-5\\  
  &AIR~\cite{eslami2016attend}            &0.106$\pm$3e-2      &0.175$\pm$4e-2 &0.044$\pm$4e-2
  &0.052$\pm$2e-2  &0.123$\pm$7e-3      &4.5e-3$\pm$6e-4 &0.211$\pm$1e-2 \\ 
  &GMIOO~\cite{yuan2019generative}         &0.463$\pm$2e-2   &0.859$\pm$2e-2  &0.228$\pm$1e-2
  &0.834$\pm$4e-2      &0.468$\pm$3e-2      & \bfseries 3.2e-3$\pm$4e-4 &0.149$\pm$1e-2 \\ 
  &SPACE~\cite{lin2019space}       &0.311$\pm$2e-2      &0.758$\pm$3e-2   &0.068$\pm$2e-2
  & 0.605$\pm$5e-2      &0.407$\pm$3e-2      &3.9e-3$\pm$1e-3 &0.136$\pm$1e-3  \\ 
  &GENESIS~\cite{engelcke2019genesis}         &     0.217$\pm$1e-6      &     0.607$\pm$5e-6       &    0.058$\pm$2e-7
  &     0.463$\pm$6e-6      &     0.204$\pm$1e-6      &     4.7e-3$\pm$2e-9 & 0.141$\pm$4e-7\\ 
  &GENESIS-V2~\cite{engelcke2021genesis}    & 0.574$\pm$2e-2  &0.894$\pm$2e-2  &0.601$\pm$2e-2
  &0.893$\pm$3e-2      &0.531$\pm$2e-2      &7.0e-3$\pm$1e-3 &0.199$\pm$1e-2 \\ 
  &Slot Attention~\cite{locatello2020object} & \bfseries 0.610$\pm$3e-4   &0.807$\pm$3e-4   & \bfseries 0.694$\pm$4e-4
  &0.738$\pm$7e-4      &0.536$\pm$9e-4     &7.4e-3$\pm$2e-5  &0.249$\pm$1e-4  \\ 
  &EfficientMORL~\cite{emami2021efficient}   & 0.279$\pm$2e-4      &  0.553$\pm$8e-4    & 0.113$\pm$1e-4
  & 0.279$\pm$2e-4      &  0.189$\pm$3e-4      &  1.1e-2$\pm$1e-5 & 0.409$\pm$2e-4\\ 
  &SLATE~\cite{singh2022illiterate}           & 0.219$\pm$2e-4      &  0.653$\pm$6e-4       & 0.039$\pm$9e-5
  & 0.476$\pm$9e-4      &  0.163$\pm$2e-4      & 2.6e-2$\pm$2e-5 & 0.283$\pm$4e-4 \\ 
  &BO-QSA (mix)~\cite{Jia2022ImprovingOL} &0.583$\pm$5e-8   & \bfseries 0.901$\pm$8e-8   &  0.354$\pm$4e-8
  & \bfseries 0.913$\pm$3e-8      & \bfseries 0.662$\pm$1e-7     &3.7e-3$\pm$1e-8  & 0.150$\pm$2e-7 \\ 
  &BO-QSA (trans)~\cite{Jia2022ImprovingOL} &0.479$\pm$6e-7   &0.821$\pm$1e-6   &    0.204$\pm$4e-7
  &0.823$\pm$1e-6      &0.573$\pm$2e-6     &4.5e-3$\pm$1e-6  & \bfseries 0.117$\pm$7e-5 \\
  \midrule
    \multirow{2}{*}{Video-based} 
  &SAVi~\cite{kipf2022conditional}           & 0.362$\pm$1e-4      & \bfseries 0.915$\pm$4e-4       & 0.099$\pm$1e-4
  & \bfseries 0.916$\pm$6e-4      &  0.467$\pm$1e-3      & \bfseries 3.2e-3$\pm$2e-5 & \bfseries 0.132$\pm$3e-4 \\ 
  &STEVE~\cite{singh2022simple}         & \bfseries    0.391$\pm$4e-4      &     0.630$\pm$1e-3       &  \bfseries  0.157$\pm$1e-4
  &     0.466$\pm$1e-3      &  \bfseries   0.468$\pm$1e-3      &     8.9e-3$\pm$3e-5 &  0.143$\pm$3e-4 \\ 
  \midrule
    \multirow{2}{*}{Multi-view-based} 
  &SIMONe~\cite{kabra2021simone}         &     0.336$\pm$2e-5      &     0.634$\pm$2e-5       &    0.138$\pm$3e-5
  &     0.536$\pm$3e-5      &   \bfseries  0.271$\pm$1e-5      &     1.1e-2$\pm$3e-7 & 0.322$\pm$5e-5\\ 
  &OCLOC~\cite{yuan2022unsupervised}          & \bfseries 0.373$\pm$1e-3      & \bfseries 0.807$\pm$4e-3       & \bfseries 0.190$\pm$4e-4
  & \bfseries 0.799$\pm$6e-3      & 0.014$\pm$4e-4      & \bfseries 9.0e-3$\pm$9e-5 & \bfseries 0.252$\pm$9e-4\\
  \bottomrule 
  
  \end{tabular}
  }
\end{table*}

\paragraph{Background segmentation performance} Although several methods have good object segmentation performance, most of these methods failed to segment the whole scene, as shown by AMI-A and ARI-A. We attribute this result to the fact that they are more likely to segment the large background into several parts (since the background in the proposed dataset is not a solid color) to ensure that they can learn more than one representation to represent the complex background. On the contrary, some methods, such as GENESIS-V2~\cite{engelcke2021genesis}, group the background into a single cluster so that they can cleanly separate the foreground objects from the background. In addition, Methods such as GMIOO~\cite{yuan2019generative}, SPACE~\cite{lin2019space}, and OCLOC~\cite{yuan2022unsupervised}, which model the foreground and background separately, generally outperform methods that do not model backgrounds in background segmentation.

\paragraph{Scene reconstruction performance} Although these methods can reconstruct scenes well on synthesis datasets, which makes researchers mainly compare segmentation performance, they fail in scene reconstruction on the proposed dataset, as shown in Figure~\ref{fig:qua}. The reconstructed image typically suffers from blurriness and imprecision, with all approaches struggling to faithfully capture object intricacies such as the reflective shine on the vase's surface. In addition, smaller or occluded objects may be missing in the reconstructed image, such as the small white sphere in the example image of OCTScenes-B. Multi-view-based methods cannot reconstruct the scene as well as single-image-based methods, mainly because the consistent viewpoint-independent representations of the same objects in different viewpoints limit the ability to reconstruct different appearances. Furthermore, the two reconstruction metrics focus on different features of the image, with MSE being a pixel-level metric and LPIPS being a feature-level metric, so they are sometimes inconsistent. SLATE~\cite{singh2022illiterate} is an example of this, which performs worst in terms of MSE metrics, but its LPIPS metrics are not, since it is based on an autoregressive transformer decoder rather than a pixel-mixture decoder.

\subsection{Additional analyses}

\paragraph{Mixture-based vs. transformer-based decoder} Comparing mixture-based decoder methods, including Slot Attention BO-QSA (mix), and SAVi, with transformer-based decoder methods, including SLATE, BO-QSA (trans), and STEVE, we observe that the former have superior segmentation performance, which may be due to the simplicity of our dataset. In terms of scene reconstruction, the mixture-based decoder method has a lower MSE, but the LPIPS of the two types of methods are not significantly different, and sometimes the transformer-based decoder method is even better. This is because the pixels of the image generated by the autoregressive transformer decoder are interdependent, resulting in a degree of randomness and significant pixel-level differences from the original image. However, the autoregressive transformer decoder can capture global semantic consistency, so it tends to generate images that are more semantically consistent with the original image, resulting in low LPIPS metrics.

\begin{figure}[htb]
  \centering
  \includegraphics[width=0.95\columnwidth]{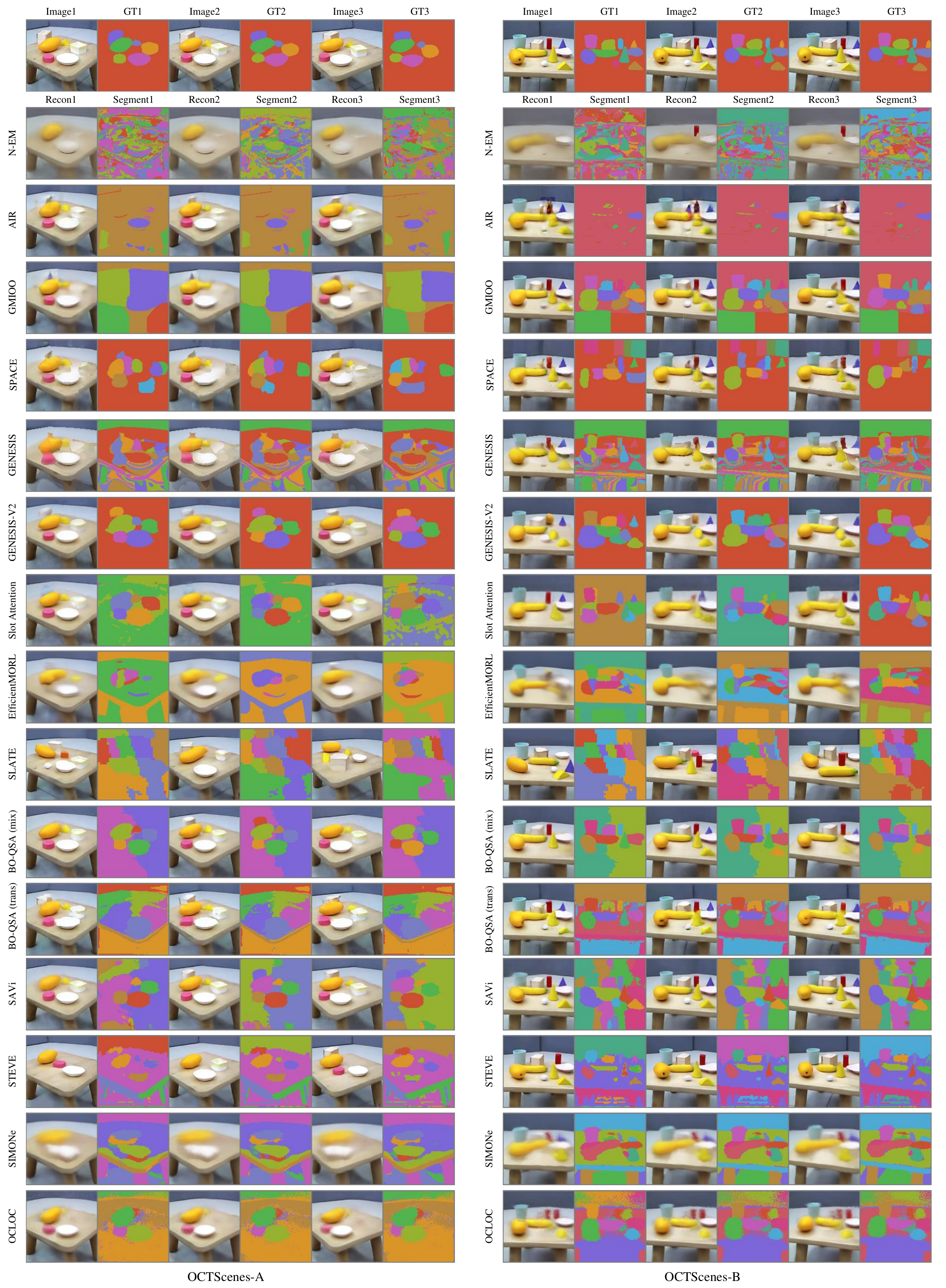}
  \caption{Qualitative results of the representative object-centric learning methods on OCTScenes-A and OCTScenes-B datasets.}
  \label{fig:qua}
\end{figure}

\paragraph{Slot Attention vs. BO-QSA} Although SLATE, STEVE, and BO-QSA (trans) are all based on transformer decoders, their performance varies greatly. Both SLATE and STEVE employ the original Slot Attention module, and their performance is highly unstable, relying heavily on slot initialization sampled from Gaussian distributions. While BO-QSA (trans) has more stable model performance, i.e. smaller variances in results, across different trials of experiments. This is because BO-QSA directly learns slot initialization as a query instead of sampling from learnable Gaussian distributions, thereby reducing the randomness of the model and improving segmentation performance. Therefore, BO-QSA may be more suitable for transformer-based decoder methods than Slot Attention.

\textbf{Results on OCTScenes-A vs. OCTScenes-B} The results show that the segmentation and reconstruction performance of the various methods on OCTScenes-A is almost identical to that on OCTScenes-B, even though the number of objects and object types increases in OCTScenes-B, which means more occlusions between objects, as well as the number of slots. Remarkably, the segmentation performance of these methods on OCTScenes-B is comparable to and even better than, that on OCTScenes-A. In terms of reconstruction performance, most methods have larger MSE on OCTScenes-B, but smaller LPIPS. The results indicate that object-centric learning methods have strong scalability with respect to the number and variety of objects, making them well-suited for adapting to real-world scenes with a richer variety of objects. The observed improvements in both segmentation and feature-level reconstruction performance could be attributed to the greater abundance of training images in the OCTScenes-B dataset compared to OCTScenes-A. The increase in the amount of training samples may allow the models to better capture the underlying patterns and complexities present in the scenes, resulting in more robust and accurate performance results. On the other hand, the decrease in pixel-level reconstruction performance may be due to the more complex and severe occlusion in OCTScenes-B, making it difficult to accurately reconstruct pixel-level details.

\section{Conclusions}
This paper proposes a versatile real-world dataset of tabletop scenes for object-centric learning, called OCTScenes, to evaluate and analyze object-centric learning methods as a benchmark dataset to fill the lack of real-world scene datasets. OCTScenes contains 5,000 scenes with a total of 15 objects in 60 frames covering a 360-degree perspective. As a result, OCTScenes can simultaneously satisfy single-image, video, and multi-view methods. Extensive experiments show that the OCTScenes dataset is suitable for evaluating object-centric learning methods. The results show that the proposed dataset is quite challenging for existing methods, illustrating the importance of OCTScenes for the research and development of object-centric learning methods.

\paragraph{Limitations} The main limitation of the dataset is its simplicity, characterized by a single background type and uncomplicated object shapes, most of which are symmetrical and lack the variation in orientation that occurs when viewed from different perspectives. Therefore, the object representations learned by the model are relatively simple, and some simple modeling methods may produce better segmentation results than complex modeling methods. 

\paragraph{Future work} To overcome the aforementioned issue and enhance the dataset further, we have devised a plan for the next version of OCTScenes. One of the primary improvements we intend to make is introducing a wider array of diverse and complex backgrounds, encompassing tables with varying types, patterns, and materials. This will allow us to simulate a multitude of real-world tabletop scenes, creating a more authentic setting for learning object-centric representation. Additionally, we recognize the need to introduce a greater variety of objects into the OCTScenes, particularly objects with asymmetrical shapes, complex textures, and mixed colors. By diversifying the object pool to include more complex objects, models can effectively capture the intricacies and nuances of object-centric representation. This expansion will not only enrich the dataset but also enable object-centric learning methods to explore a broader spectrum of object properties, such as shape, texture, and color. As a result, it will facilitate more comprehensive learning and evaluation. 

In summary, the upcoming version of OCTScenes will address the limitations of the current dataset by introducing more complex backgrounds and a wider variety of objects. These enhancements will propel object-centric learning forward, allowing researchers to delve deeper into the complexities of visual perception and object understanding.


\bibliographystyle{IEEEtran}
\bibliography{reference}

\newpage

\appendix

\section{Detailed dataset description}

\subsection{Dataset construction}
\label{dataset}
The dataset can be accessed at \url{https://huggingface.co/datasets/Yinxuan/OCTScenes}. OCTScenes is available under CC-BY-NC 4.0 license. In OCTScenes, each instance contains 60 frames of RGB-D images depicting a tabletop scene, captured from multiple viewpoints. Each image is available in three different sizes: 640$\times$480, 256$\times$256, and 128$\times$128, along with their corresponding depth maps and segmentation maps. All images are provided as PNG.

Before being inputted into the models, the raw data with a resolution of 640$\times$480 underwent a series of pre-processing steps. Firstly, it was center-cropped based on the manually labeled bounding box of the table, resulting in a 256$\times$256 patch. Subsequently, the cropped image was further down-sampled to 128$\times$128 pixels. This process removes uninteresting empty edges of the scenes and reduces the computational load. Many of the benchmarked models were developed to work with such resolution. For convenience, we provide the relevant code for data processing on our website.

\section{Experimental details}
\label{experiment}
All benchmarked models were trained on NVIDIA GeForce 3090 GPUs. To fully utilize the images for training, we divided each scene with 60 frames into 6 sub-scenes with an interval of 10 frames and sampled 3 scenes from the 6 sub-scenes at intervals. For single-image-based methods, all the frames in a scene were used as input. For video-based methods, we divided every 3 consecutive frames in a scene as a training sample, resulting in a total of 3 training samples per scene. For multi-view-based methods, we randomly selected 4 frames in a scene as a training sample, resulting in only one training sample per scene. The number of slots was set to 8 for OCTScenes-A, while it was set to 12 for OCTScenes-B. All the reported results are based on 3 evaluations of the testing sets.

\subsection{Hyperparameters}

\paragraph{N-EM~\cite{greff2017neural}} We used the unofficial N-EM implementation in the toolbox of compositional scene representation\footnote{\url{https://github.com/FudanVI/compositional-scene-representation-toolbox}}. Models were trained with the default hyperparameters described in the "experiments\_benchmark/config\_clevr.yaml" file of the code repository, with the exception of the batch size, which was set to 64.

\paragraph{AIR~\cite{eslami2016attend}} We used the unofficial AIR implementation in the toolbox of compositional scene representation. Models were trained with the default hyperparameters described in the "experiments\_benchmark/config\_clevr.yaml" file of the code repository, with the exception of the batch size, which was set to 64.

\paragraph{GMIOO~\cite{yuan2019generative}} We used the official GMIOO implementation in the toolbox of compositional scene representation. Models were trained with the default hyperparameters described in the "experiments\_benchmark/config\_clevr.yaml" file of the code repository, with the exception of the batch size, which was set to 64.

\paragraph{SPACE~\cite{lin2019space}} We used the official SPACE implementation in the toolbox of compositional scene representation. Models were trained with the default hyperparameters described in the "src/configs/clevr.yaml" file of the code repository, with the exception of the batch size, which was set to 64.

\paragraph{GENESIS~\cite{engelcke2019genesis}} We used the unofficial GENESIS implementation in the toolbox of compositional scene representation. Models were trained with the default hyperparameters described in the "genesis/models/genesis\_config.py" file of the code repository, with the exception of the batch size, which was set to 64.

\paragraph{GENESIS-V2~\cite{engelcke2021genesis}} We modified the unofficial GENESIS-V2 implementation in the toolbox of compositional scene representation. Models were trained with the default hyperparameters described in the "genesis/models/genesisv2\_config.py" file of the code repository, with the exception of the batch size, which was set to 64.

\paragraph{Slot Attention~\cite{locatello2020object}} We used the unofficial Slot Attention implementation in the toolbox of compositional scene representation. Models were trained with the default hyperparameters described in the "experiments\_benchmark/config\_clevr.yaml" file of the code repository.

\paragraph{EfficientMORL~\cite{emami2021efficient}} The official EfficientMORL implementation\footnote{\url{https://github.com/pemami4911/EfficientMORL}} was used. Models were trained with the default hyperparameters described in the "configs/train/clevr6-128x128/EMORL.json" file of the code repository, with the exception of the batch size, which was set to 64.

\paragraph{SLATE~\cite{singh2022illiterate}} The official SLATE implementation\footnote{\url{https://github.com/singhgautam/slate}} was used. The hyperparameters were similar to the ones described in the original SLATE paper for CLEVRTex, with the exception of the batch size, which was set to 64.

\paragraph{BO-QSA~\cite{Jia2022ImprovingOL}} The official BO-QSA implementation\footnote{\url{https://github.com/YuLiu-LY/BO-QSA}} was used. The hyperparameters for the mixture-based decoder were similar to the ones described in the "scripts/train/mix\_dec\_clevrtex.sh" file of the code repository, with the exception of the batch size, which was set to 64. The hyperparameters for the transformer-based decoder were similar to the ones described in the "scripts/train/trans\_dec\_coco.sh" file of the code repository, with the exception of the batch size, which was set to 64.

\paragraph{SAVi~\cite{kipf2022conditional}} We modified the official SAVi implementation\footnote{\url{https://github.com/google-research/slot-attention-video}} into a PyTorch version, and used the unsupervised version trained only with videos. The architecture and hyperparameters closely followed the original SAVi paper for the MOVi++ dataset with 128$\times$128 input resolution, except for the batch size, which was set to 8.

\paragraph{STEVE~\cite{singh2022simple}} The official STEVE implementation\footnote{\url{https://github.com/singhgautam/steve}} was used. Models were trained with the default hyperparameters described in the "train.py" file of the official code repository, except for the batch size, which was set to 8.

\paragraph{SIMONe~\cite{kabra2021simone}} We modified the unoffcial SIMONe implementation\footnote{\url{https://github.com/lkhphuc/simone}} into a PyTorch version. The architecture and hyperparameters closely followed the original SIMONe paper with the following differences: 1)the batch size was set to 8; 2)if the number of slots is 8, the feature map computed by the CNN of the encoder will be input into a sum pooling layer with kernel size (4,2) and stride (4,2); 3)if the number of slots is 12, the feature map computed by the CNN of the encoder will be input into a max pooling layer with kernel size (3,2) and stride (2,2).

\paragraph{OCLOC~\cite{yuan2022unsupervised}} The official OCLOC implementation in the toolbox of compositional scene representation was used. Models were trained with the default hyperparameters described in the "exp\_multi/config\_shop\_multi.yaml" file of the official code repository, except for the batch size, which was set to 8.

\section{Additional results}
\label{results}
\begin{figure}[p]
  \centering
  \includegraphics[width=\columnwidth]{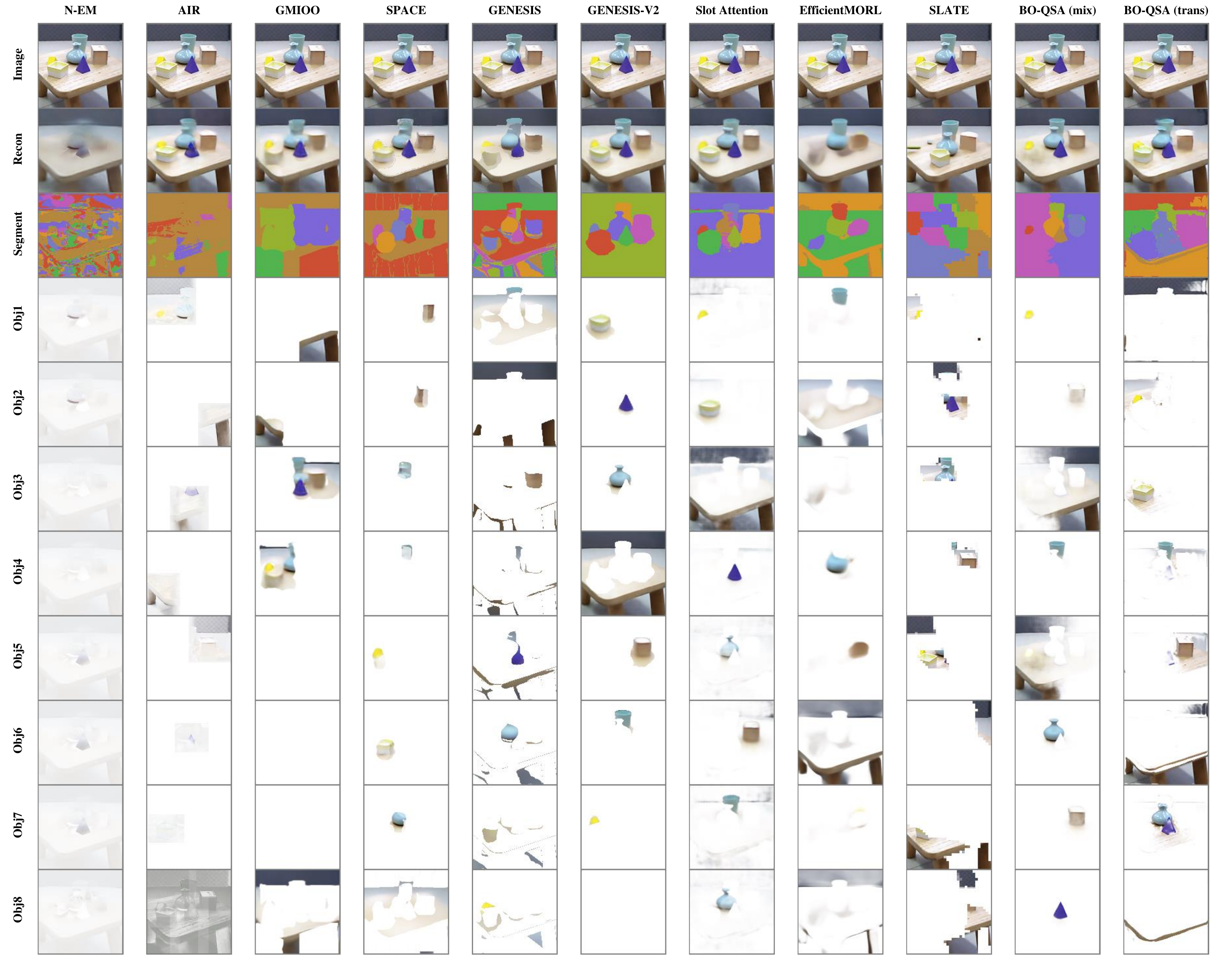}
  \caption{Qualitative results of the representative object-centric learning methods based on single-image on OCTScenes-A dataset.}
  \label{fig:static-a}
\end{figure}

\begin{figure}[p]
  \centering
  \includegraphics[width=\columnwidth]{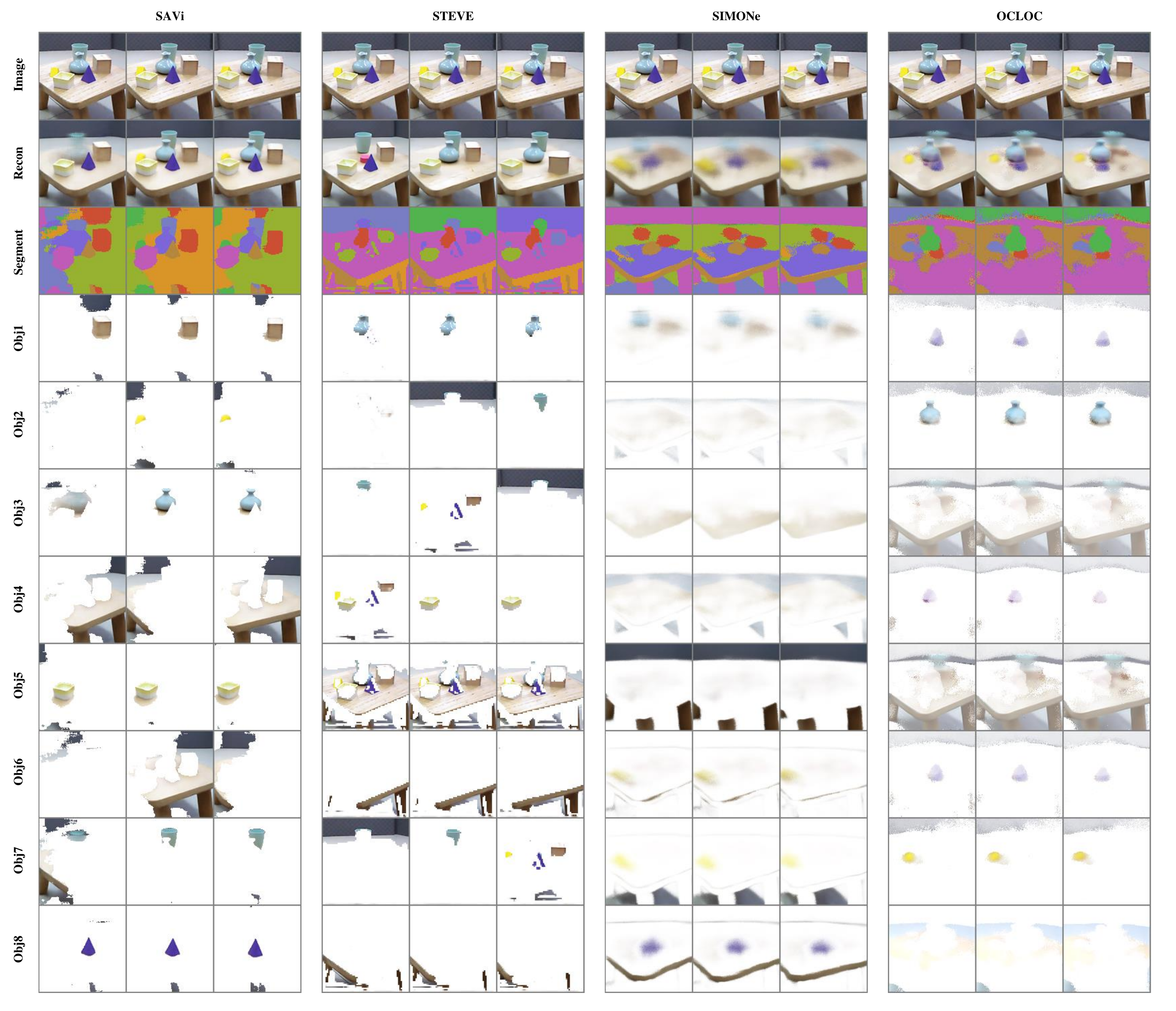}
  \caption{Qualitative results of the representative object-centric learning methods for dynamic scenes and multi-view scenes on OCTScenes-A dataset.}
  \label{fig:multi-a}
\end{figure}

\begin{figure}[p]
  \centering
  \includegraphics[width=\columnwidth]{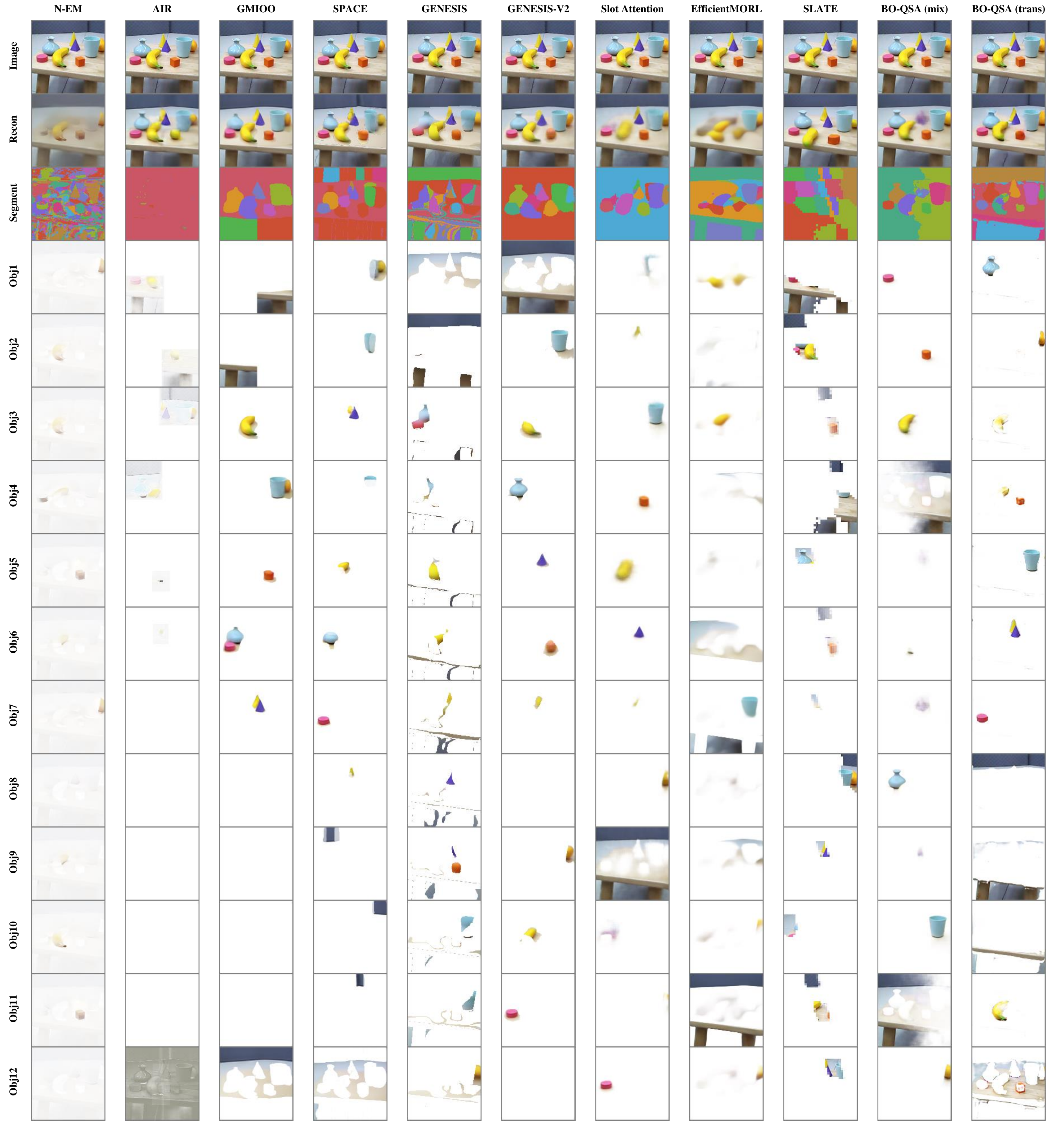}
  \caption{Qualitative results of the representative object-centric learning methods based on single-image on OCTScenes-B dataset.}
  \label{fig:static-b}
\end{figure}

\begin{figure}[p]
  \centering
  \includegraphics[width=\columnwidth]{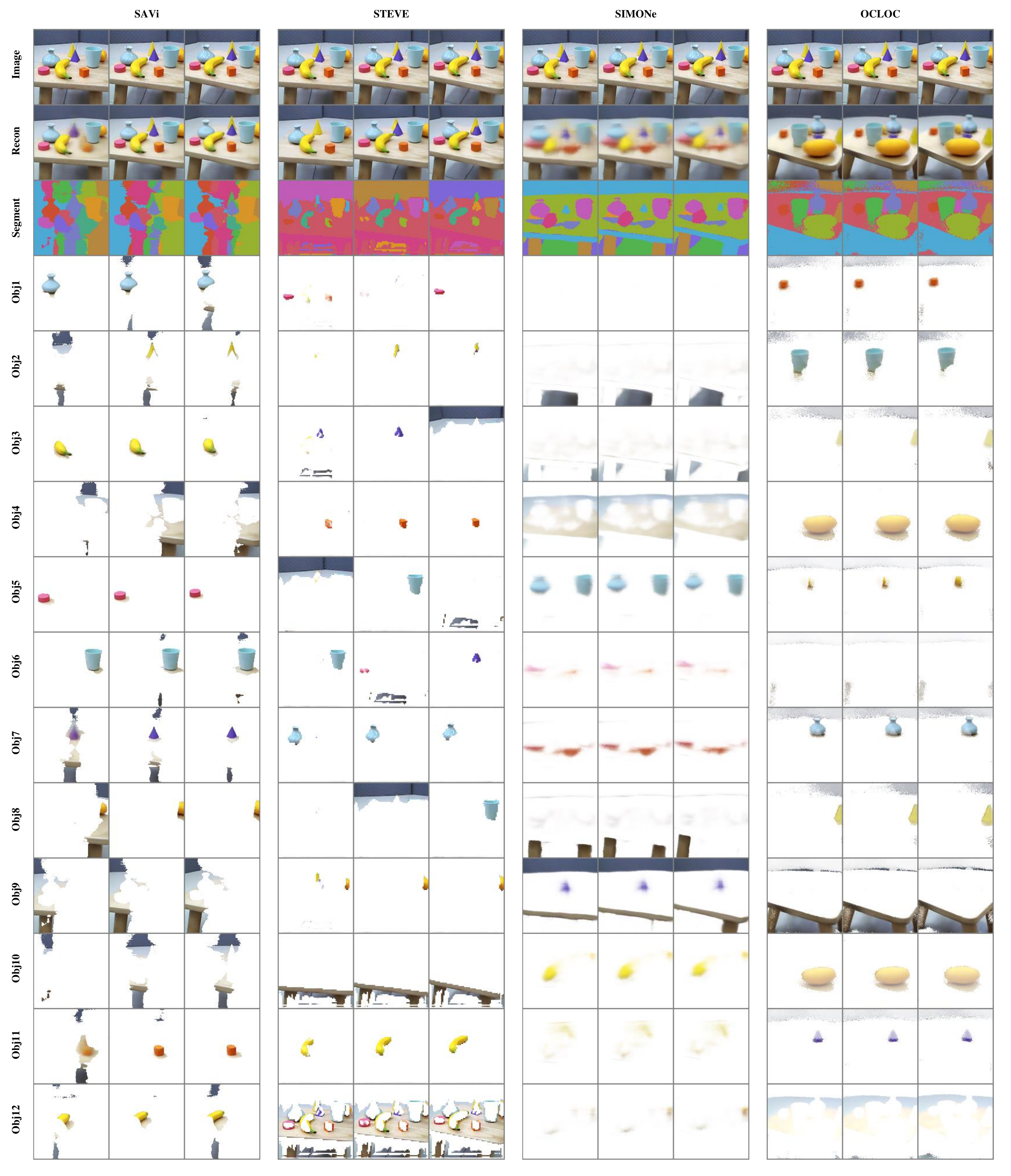}
  \caption{Qualitative results of the representative object-centric learning methods for dynamic scenes and multi-view scenes on OCTScenes-B dataset.}
  \label{fig:multi-b}
\end{figure}

We present additional qualitative scene decompositions for all benchmarked models on OCTScenes-A in Figure~\ref{fig:static-a} and Figure~\ref{fig:multi-a} and OCTScenes-B in Figure~\ref{fig:static-b} and Figure~\ref{fig:multi-b}.

\paragraph{Segmentation performance} In terms of model segmentation performance, some methods such as GENESIS-V2~\cite{engelcke2021genesis}, Slot Attention~\cite{locatello2020object}, BO-QSA~\cite{Jia2022ImprovingOL} and SAVi~\cite{kipf2022conditional} have demonstrated the ability to decompose scenes into meaningful individual objects. While some methods segment small or occluded objects as backgrounds, such as EfficientMORL~\cite{emami2021efficient}, or split multiple objects into the same object, such as GMIOO~\cite{yuan2019generative}. From the visualization results, it can be seen that the objects segmented by STEVE~\cite{singh2022simple} are incomplete, and the edges of the objects are divided into backgrounds. Therefore, the obtained object mask is often smaller than the ground truth, and the corresponding segmentation metrics are lower. This may be because the mask used by STEVE is the result of up-sampling the attention mask, which is not as accurate as the mask generated by the mixture-based decoder.

\paragraph{Reconstruction performance} Most methods can reconstruct images with a high degree of similarity to the original image, but they are often blurry and may miss some small or occluded objects. It is worth noting that the vast majority of methods cannot reconstruct scene details. For example, only SLATE and STEVE can reconstruct the shine on the surface of the vase, while other methods cannot.

In summary, OCTScenes poses significant challenges for existing methods, therefore it can promote innovation and improvement of object-centric learning methods.

\end{document}